\documentclass[letterpaper, 10 pt, conference]{ieeeconf}  
\IEEEoverridecommandlockouts                              

\usepackage{amsmath} 
\usepackage{amssymb} 
\usepackage{float}
\usepackage[pdftex]{graphicx}
\usepackage{tabularx}
\usepackage{verbatim}
\usepackage{color}
\usepackage{xparse}
\usepackage[hidelinks]{hyperref}   
\usepackage{xmpmulti}
\usepackage{transparent}
\usepackage{fancyhdr}
\usepackage{cite}
\usepackage{rotating}
\usepackage{bbm}
\usepackage{amssymb,amsmath,float}
\usepackage[linesnumbered, algoruled, vlined]{algorithm2e}
\usepackage{booktabs}
\usepackage{caption}
\usepackage{subcaption}
\usepackage{tikz}
 
\usepackage{amsthm}
\usepackage{hyperref}
\usepackage{wrapfig}
\usepackage{setspace}
\usepackage{graphicx}
\usepackage{epsfig}
\usepackage[all]{xy}
\usepackage{verbatim}
\usepackage{float}
\usepackage{booktabs}
\usepackage{multirow}
\usepackage{color,soul}
\usepackage{amssymb,amsmath,graphicx}
\usepackage[compact]{titlesec}
\usepackage{diagbox}
\usepackage{array}
\usepackage{units}
\usepackage[export]{adjustbox}

\DeclareMathOperator*{\argmax}{arg\,max}

\usepackage{dsfont}

\title{\LARGE \bf
Learning Nash Equilibrial Hamiltonian for Two-Player Collision-Avoiding Interactions}

\author{Lei Zhang$^{1}$, Siddharth Das$^{1}$, Tanner Merry$^{1}$, Wenlong Zhang$^{2}$, Yi Ren$^{1}$ 
\thanks{$^{1}$L. Zhang, S. Das, T. Merry, and Y. Ren are with the School of Engineering, Matter, Transport, and Energy, Arizona State University, Tempe, AZ 85287, USA. Email:
        {\tt\small \{lzhan300, sdas100, tmerry, yiren\}@asu.edu}}%
\thanks{$^{2}$W. Zhang is with the School of Manufacturing Systems and Networks, Arizona State University, Mesa, AZ 85212, USA. Email:
        {\tt\small wenlong.zhang@asu.edu}}%
}

\begin{document}

\maketitle
\thispagestyle{empty}
\pagestyle{empty}

\begin{abstract}
We consider the problem of learning Nash equilibrial policies for two-player risk-sensitive collision-avoiding interactions. Solving the Hamilton-Jacobi-Isaacs equations of such general-sum differential games in real time is an open challenge due to the discontinuity of equilibrium values on the state space. A common solution is to learn a neural network that approximates the equilibrium Hamiltonian for given system states and actions. The learning, however, is usually supervised and requires a large amount of sample equilibrium policies from different initial states in order to mitigate the risks of collisions. This paper claims two contributions towards more data-efficient learning of equilibrium policies: First, instead of computing Hamiltonian through a value network, we show that the equilibrium co-states have simple structures when collision avoidance dominates the agents' loss functions and system dynamics is linear, and therefore are more data-efficient to learn. Second, we introduce theory-driven active learning to guide data sampling, where the acquisition function measures the compliance of the predicted co-states to Pontryagin's Maximum Principle. On an uncontrolled intersection case, the proposed method leads to more generalizable approximation of the equilibrium policies, and in turn, lower collision probabilities, than the state-of-the-art under the same data acquisition budget.   
\end{abstract}

\vspace{5pt}
\section{Introduction}
Physical interactions between intelligent agents (e.g., humans and robots) can be considered as general-sum differential games where the equilibrium policies are governed by the Hamilton-Jacobi-Isaacs (HJI) equations~\cite{evans1984differential}. 
Such interactions are ubiquitous in transportation, health care, and defense applications, and fast computation of equilibrium policies is critical for real-time intent inference~\cite{chen2020shall}, motion planning, and signaling~\cite{brown2020combining} that are necessary for interactions in social environments.
However, solving HJI equations in real time is still challenging for risk-sensitive interactions~\cite{zhang2024value}, where collision avoidance induces discontinuity in agents' equilibrium values and policies, which are functions defined on the space of system states.

Existing solutions can be categorized into learning, control, or a mixture of these two approaches: Learning approaches, e.g., variants of multi-agent reinforcement learning (MARL), approximate policies by converging both the dual (i.e., the values) and the primal (i.e., the policies) variables. However, the state-of-the-art model-free MARL, e.g., counterfactual regret minimization (CFR)~\cite{zinkevich2007regret} and fictitious self-play (FP)~\cite{heinrich2015fictitious}, still have difficulty to converge in games with large state/action spaces or large depth~\cite{brown2019deep,brown2020combining}. 
Some control methods, on the other hand, find policies for given initial states using Pontryagin's Maximum Principle (PMP)~\cite{pontryagin1966theory}, and iteratively do so in real time~\cite{fridovich2020efficient,leung2020infusing}. To enable fast computation, these methods leverage known system dynamics and solve depth-limited sub-games, e.g., via model predictive control. While methods for solving HJI equations exist, e.g., through level set~\cite{osher1991high,mitchell2005time}, these are not scalable to problems with more than a handful of state dimensions.
Our method belongs to the third category, where equilibrium policies are sampled over the set of initial states and solved through PMP, and then used for learning a policy approximation in a supervised fashion~\cite{nakamura2019adaptive,zhang2023approximating}. This approach leverages knowledge about system dynamics while generalizing solutions through neural networks.

Compared with static supervised learning, this paper improves the generalization performance of learned policies through the following contributions:
\begin{itemize}
    \item First, we show that the equilibrium co-states are intrinsically low-dimensional \textit{when system dynamics are linear}, and are data efficient to learn. Since Hamiltonian can be computed based on the co-state, the system dynamics, and the loss functions, this finding leads to a data-efficient way of computing equivalent Hamiltonian, from where the policy can be derived. We empirically validate this result on a two-player uncontrolled intersection case.  
    
    \item Second, we introduce theory-driven active learning to guide data sampling. Given an initial system state, the acquisition function measures the compliance of the predicted co-states to PMP. Since computing the compliance involves rolling out the system dynamics backward only once, which is much cheaper than solving the BVP, the proposed active learning algorithm leads to efficient sampling of initial system states for which the co-state predictions have large errors.
\end{itemize}

\vspace{5pt}
\section{Related Work}
\label{sec:related}

\textbf{Value approximation}: Value approximation, i.e., approximating dual solutions to HJB or HJI equations, has been studied under model-free and model-based settings and in contexts of optimal control~\cite{borovykh2022data}, zero-sum differential games~\cite{bansal2021deepreach}, and general-sum differential games~\cite{zhang2023approximating}. Our method follows a common strategy where we approximate the action-value of policies using a neural network, which is trained on sampled policies for individual initial states of the games. These state-specific policies are derived through standard solvers, e.g., a BVP solver for optimal control~\cite{nakamura2019adaptive}, and CFR, FP, or Monte Carlo tree search for depth-limited stochastic games~\cite{brown2020combining}. Our study is unique in that instead of learning action-values directly, we learn co-states, i.e., the value gradients with respect to states, and leverage known system dynamics and loss functions to derive the action-values (or Hamiltonian in differential games).

\textbf{Physics-based learning}: Our method is also related to existing studies that learn system dynamics through Hamiltonian. For example, \cite{zhong2019symplectic} used Hamilton's equation of motion to learn Hamiltonian from sample trajectories of dynamical systems. Instead of considering Hamiltonian as a blackbox, our method learns the generalized momentum of the system (i.e., the co-states), from where the Hamiltonian can be derived.

\textbf{Self-supervised learning (SSL)}: SSL considers the existence of governing equations in the form of $h(x,y(x))=0$ for input $x$ and a predictive model $y$ of $x$~\cite{jing2020self,misra2020self}. These equations constrain the learning of a mapping between $x$ and $y$. In computer vision, for example, $h$ may represent the invariance of features $y$ with respect to transformations of image $x$~\cite{misra2020self}. In this study, $h$ represents the boundary-value problem (BVP) derived from PMP, $x$ the initial system states, and $y$ the predicted equilibrium co-states. A difference from SSL is that, instead of learning $y$ completely based on $h$, which would be equivalent to solving HJI equations directly and is hard, we use $h$ to guide the adaptive sampling of $(x,y)$ using an existing BVP solver.

\textbf{Active learning}: When labels are costly for supervised learning, one may actively choose data for which the labels are expected to most effectively improve the generalization of the learned model. Traditional acquisition functions, i.e., goodness metrics that determine which data to sample, are constructed based on statistical uncertainties in prediction~\cite{ren2021survey}. These methods are agnostic to the underlying governing equations $h(x,y)=0$. As a different approach to active learning, \cite{cang2019one} demonstrated the efficacy of a theory-driven acquisition function that measures the compliance of predicted optimal solutions to the KKT conditions. Our method extends \cite{cang2019one} to the setting of differential games.

\vspace{5pt}
\section{Problem Statement}
We consider games where both players share the same individual action set $\mathcal{U}$ and state space $\mathcal{X}$. Let $a_i$ and $a_{-i}$ be placeholder variables for player $i$ and its fellow player, respectively. We use bold symbols, e.g., $\textbf{a} = (a_1,a_2)$, to represent variable pairs. Each player $i$ has an instantaneous loss function $l_i(\textbf{x},u_i)$, a terminal loss function $g_i(x_i)$, and a time-invariant dynamical model $\dot{x}_i = f_i(x_i,u_i)$. Let the finite time horizon be $[0,T]$. A differential game with initial state $\textbf{x}_0$ and time $t_0$ is denoted by $\mathcal{G}(\textbf{x}_0, t_0) := <\mathcal{X}, \mathcal{U}, \textbf{l}, \textbf{g}, \textbf{f}, T, \textbf{x}_0, t_0>$, where $\textbf{l}, \textbf{g}, \textbf{f}$ are the abbreviations for instantaneous loss, terminal loss and dynamical model. 
The Hamiltonian is defined as
\begin{equation}
    H_i(u_i; \textbf{x}, \lambda_i, t) = \lambda_i^T\textbf{f} - l_i(\textbf{x},u_i),
\label{eq:hamiltonian}
\end{equation}
where $\lambda_i$ is the co-state. Let Nash equilibrium be the solution concept for non-cooperative players and given $(\textbf{x}_0, t_0)$, the open-loop equilibrium policies of $\mathcal{G}(\textbf{x}_0, t_0)$ follow PMP:
\begin{equation}
\begin{aligned}
    & \dot{x}_i^{*} = f_i(x_i^{*}, u_i^{*}) \\
    & \textbf{x}^{*}(0) = \textbf{x}_0 \\
    & \dot{\lambda}_i^{*} = - \nabla_{x_i} H_i(u_i^{*};\textbf{x}^{*},\lambda_i^{*})\\
    & \lambda^{*}_i(T) = -\nabla_{x_i} g_i(x_i^{*}(T))\\
    & u^{*}_i = \argmax_{u_i \in \mathcal{U}} H_i(u_i;\textbf{x}^*,\lambda_i^*),
    ~\text{ for } i\in \{1,2\},
    \label{eq:pmp}
\end{aligned}
\end{equation}
where $\textbf{x}^*$, $\textbf{u}^*$, and $\boldsymbol{\lambda}^{*}$ are the equilibrium state, action, and co-state trajectories, respectively, and are dependent on $\textbf{x}_0$ and $t_0$. 
Eq.~\eqref{eq:pmp} represents a boundary value problem (BVP) where $\textbf{x}$ has initial conditions at $t=t_0$ while $\boldsymbol{\lambda}$ has boundary condition at $t=T$. Solving Eq.~\eqref{eq:pmp} for a specific $\textbf{x}_0$ and $t_0$ can be achieved through a standard BVP solver~\cite{kierzenka2001bvp}. 
Given a set of initial states $\mathcal{X}_0$ and $\mathcal{T}_0$, our goal is to approximate the equilibrium policies that solve $\mathcal{G}(\textbf{x}, t)$ for any $\textbf{x} \in \mathcal{X}_0$ and $t \in \mathcal{T}_0$. This will be achieved by learning a co-state network $\hat{\boldsymbol{\lambda}}(\textbf{x}_0, t_0)$, from where $\textbf{u}^*$ can be computed by locally maximizing $H_1$ and $H_2$.

\vspace{10pt}
\section{Methods}
In this section, we discuss methods that improve the data efficiency of learning to approximate the equilibrium policies.

A baseline approach is to collect data in the form of $(\textbf{x}_0, t, V_i^*(\textbf{x}_0, t))$ for supervised learning, where $V_i^*(\textbf{x}_0, t) = \int_{\tau=t}^T l_i(\textbf{x}^*(\tau),u_i^*(\tau)) d\tau + g_i(x_i^*(T))$ is the equilibrium value of player $i$ starting at $\textbf{x}_0$ and time $t$. By learning to predict the value and leveraging $\lambda^*_i = \nabla_{\textbf{x}}V_i$, we can then perform closed-loop control of player $i$ via maximizing the Hamiltonian parameterized by $\lambda^*_i$. The potential drawback of this approach is that the error in value prediction can be amplified through the gradient operator $\nabla_{\textbf{x}}$, leading to larger variances in the prediction error of the co-states.

As an alternative, we propose to directly learn to approximate equilibrium co-state trajectories. This will be supported by two insights we discuss in the following: First, we show that with linear system dynamics and a collision penalty, the equilibrium co-state trajectories, while being continuous in time, have low-dimensional representations. Second, the co-states are constrained by PMP, thus its prediction error can be measured by solving a single inverse value problem (i.e., rolling back the dynamics using the predicted terminal states and co-states) without relying on the ground truth which requires solving a more expensive BVP. 

\subsection{Collision-avoiding interactions with linear dynamics}
\textbf{Co-state structure:} We model each player's instantaneous loss as
\begin{equation}
    l_i(\textbf{x},u_i) = u_i^2 + \phi_i(\textbf{x}),
    \label{eq:reward}
\end{equation}
which combines an effort loss and a collision loss $\phi_i(\cdot)$. The latter has the following properties: There exists a non-empty subset of ``collision'' states $\mathcal{C} \subset \mathcal{X} \times \mathcal{X}$, such that $\phi_i(\textbf{x}) = r$ for all $\textbf{x} \in \mathcal{C}$ and $\phi_i(\textbf{x}) =0$ otherwise, where $r$ is a large penalty. 
Let $\partial \mathcal{C}$ be the boundary of $\mathcal{C}$. $\nabla_{x_i}\phi_i(\textbf{x}) = 0$ for all $\textbf{x} \notin \partial \mathcal{C}$, and for $\textbf{x} \in \partial\mathcal{C}$, $\nabla_{x_i}\phi_i(\textbf{x})$ is a delta function. 
When system dynamics is linear, i.e., $f_i(x_i,u_i) = A_ix_i + B_iu_i$, the co-state dynamics becomes
\begin{equation}
\dot{\lambda}_i^{*}(t) = - A_i^T\lambda_i + \nabla_{x_i} \phi_i(\textbf{x}^*), ~ \lambda_i^{*}(T) = -  \nabla_{x_i} g_i(x_i^*(T)).
\label{eq:costate}
\end{equation}
While $\lambda^*_i$ is in general coupled with $\textbf{x}^*$, Eq.~\eqref{eq:costate} reveals its structure: Notice that $\lambda^*_i$ has discontinuous piece-wise linear dynamics, since for any trajectory $\textbf{x}$, $\nabla_{x_i(t)} \phi_i(\textbf{x}(t))=0$ almost everywhere on $[0,T]$ except when $x(t) \in \partial \mathcal{C}$. $\lambda^*_i$ is thus controlled by its terminal boundary $\lambda_i^{*}(T)$ and the set of time instances, denoted by $\mathcal{T}_i$, where $x(t) \in \partial \mathcal{C}$ for $t \in \mathcal{T}_i$. We argue that for two-player collision-avoiding interactions, we have the following five cases of $(\mathcal{T}_1, \mathcal{T}_2)$: $(\emptyset, \emptyset)$ (no collision), $(\{t_1,t_2\},\emptyset)$ (Player 1 overtakes Player 2), $(\emptyset,\{t_1,t_2\})$ (Player 2 overtakes Player 1), $(\{t_1\},\{t_2\})$ (Player 1 moves toward Player 2, who then evades), and $(\{t_2\},\{t_1\})$ (Player 2 moves toward Player 1, who then evades), where $0 \leq t_1 < t_2 \leq T$. Sample interactions are illustrated in Fig.~\ref{fig:discussion}a using the uncontrolled intersection case (see below). It should be noted that while categorizing different types of collisions may seem unnecessary as collisions should be avoided in the first place, softened $\phi_i$, which is necessary for correctly solving BVPs, will lead to different types of non-collision close calls (i.e., when players have small minimal distances during the interaction) where $\boldsymbol{\lambda}^*$ has similar structures as for collisions. The equilibrium values of these different types of close calls are necessary to be learned because by telling apart these interactions one may infer the types of players~\cite{chen2020shall, zhang2024value}.

\begin{figure}
    \centering
    \includegraphics[width=\linewidth]{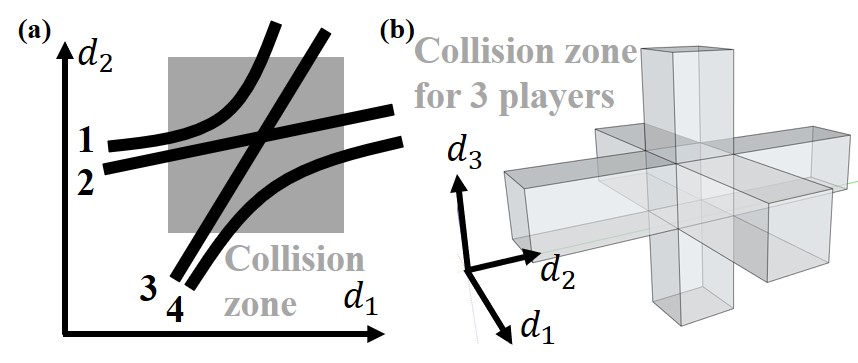}
    \caption{(a) Four types of collisions from two players at an uncontrolled intersection, characterized by the entrance and exit facets of the collision box. 1-4: Player 1 moves towards Player 2, who then evades; Player 1 overtakes Player 2; Player 2 overtakes Player 1; Player 2 moves towards Player 1, who then evades. (b) Collision zone for a three-player case, where the types of collisions (and interactions) can still be determined by the finite number of entrance and exit facets.}
    \label{fig:discussion}
\vspace{-22pt}
\end{figure}

From the above discussion, each equilibrium co-state trajectory can be expressed as
\begin{equation*}
    \lambda_i^*(t) = \exp\left(A_i^T (T-t)\right)\lambda_i^*(T) - q_i \mathds{1}(t\geq t_i^{in}) + q_i \mathds{1}(t \geq t_i^{out}),
\end{equation*}
where $t_i^{in}$ and $t_i^{out}$ are when the player moves into and out of $\mathcal{C}$, and $q_i$ is a collision-related vector. Therefore, each co-state trajectory can be summarized by 
$y_i(\textbf{x}_0)=(\lambda_i^*(T), t_i^{in}, t_i^{out}, q_i)$.

\textbf{Policy structure:} The system dynamics for Player $i$ and its fellow player follow $\textbf{f}=\left[\begin{array}{ccc}
    f_i \\
    f_{-i} \\
\end{array}\right]$, where $f_i$ and $f_{-i}$ represent linear dynamics for both players. Using Eq.~\eqref{eq:reward}, the Hamiltonian becomes
\begin{equation}
    H_i(u_i;\textbf{x},\lambda_i, t) = \lambda_i^T\textbf{f} - u_i^2 - \phi_i(\textbf{x}).
\end{equation}
Given a convex $\mathcal{U}$, $\textbf{u}^*(t)$ is thus the unique solution to a convex problem and requires low computational cost. 

\begin{figure}
    \centering
    \includegraphics[width=\linewidth]{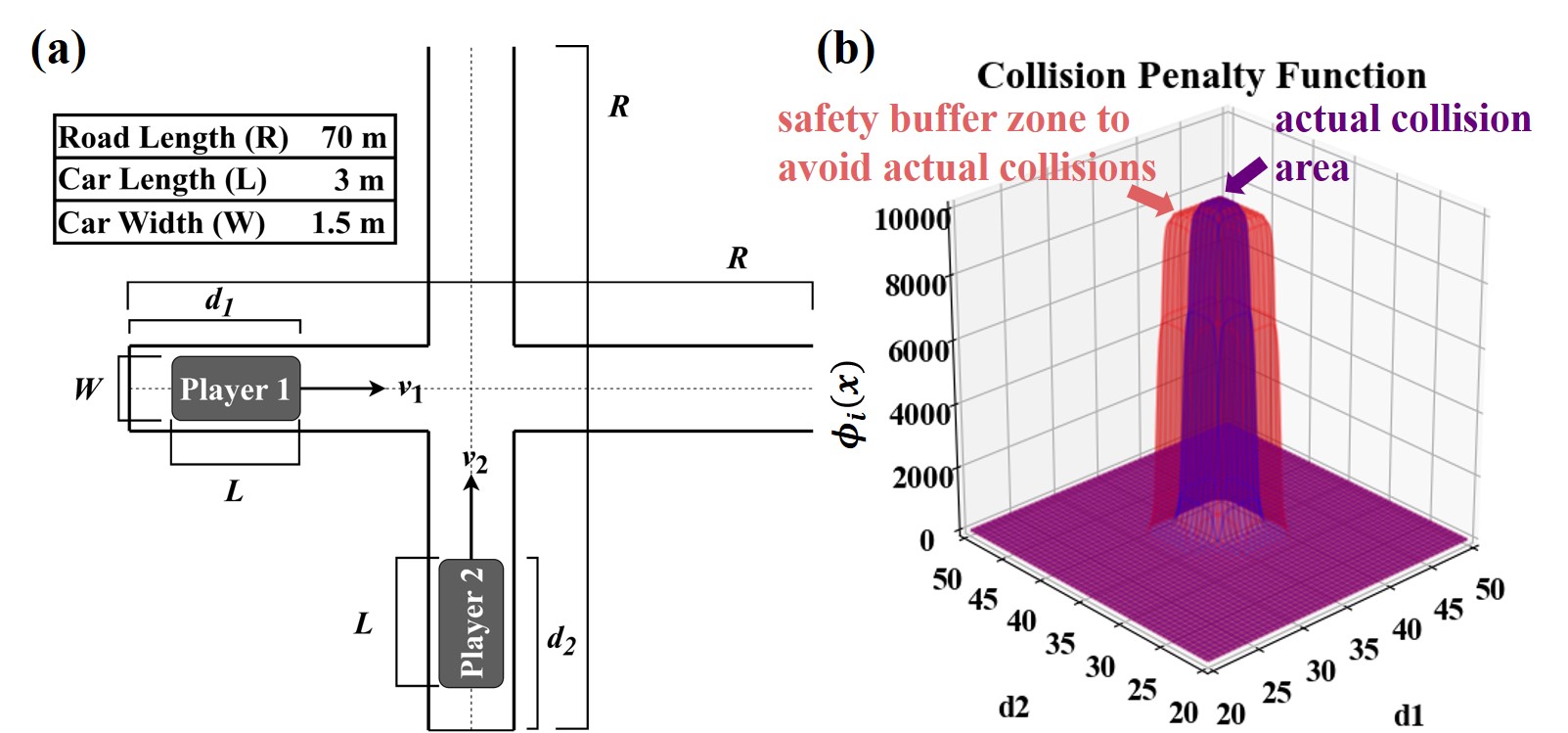}
    \caption{(a) Uncontrolled intersection setup with two players. (b) 3D visualization for collision penalty function $\phi_i(x)$: The dark purple region shows the actual collision area, while the red region represents a safety buffer to avoid collisions.}
    \label{fig:case}
\vspace{-20pt}
\end{figure}

\textbf{Running example}: We consider a two-player uncontrolled intersection case (see Fig.~\ref{fig:case}a) where each player's state is defined by its position $d_i$ and velocity $v_i$: $x_i=(d_i,v_i)$, and action by its acceleration rate. The terminal loss is $g_i(x) = -\alpha d_i(T) + (v_{i}(T)-\bar{v})^2$, which encourages the agent to move forward ($\alpha>0$) and restoring a target speed $\bar{v}$ at $T$. System dynamics follows:
\begin{equation}
    \left[
\begin{array}{ccc}
    \dot{d}_{i}(t) \\
    \dot{v}_{i}(t) \\
\end{array}
\right]
    =\left[
\begin{array}{ccc}
    0\hspace{0.4cm} 1 \\
    0\hspace{0.4cm} 0\\
\end{array}
\right]
\left[
\begin{array}{ccc}
    d_{i}(t) \\
    v_{i}(t) \\
\end{array}
\right]
+
\left[
\begin{array}{ccc}
    0 \\
    1 \\
\end{array}
\right]
u_{i}(t).
\label{eq:systemdynamics}
\end{equation}
From the above discussion, the co-state dynamics and the policy become:
\begin{equation}
\begin{aligned}
    & \dot{\lambda}_i^{*}[1] = \nabla_{d_i} \phi_i(\textbf{x}^*), ~ \lambda_i^{*}(T)[1] = \alpha\\
    & \dot{\lambda}_i^{*}[2] = -\lambda_i^{*}[1], ~\lambda_i^{*}(T)[2] = -2(v_i(T) - \bar{v})\\
    & u^*_i = 0.5\lambda^*_i[2].
\label{eq:costate_intersection}
\end{aligned}
\end{equation}
The first element of the co-state trajectory follows
\begin{equation}
    \lambda_i^*(t)[1] = \alpha - q_i \mathds{1}(t\geq t_i^{in}) + q_i \mathds{1}(t \geq t_i^{out}),
\end{equation}
while the second element $\lambda_i^*(t)[2]$ is an integral of $\lambda_i^*(t)[1]$.

\subsection{Learning equilibrium co-states}
We learn a neural network $\hat{\textbf{y}}_w(\cdot)$ that predicts $\textbf{y}(\cdot)$ for given $\textbf{x}_0 \in \mathcal{X}_0$ and time $t_0 \in \mathcal{T}_0$, where the parameters $w$ are learned through a dataset $\{(\textbf{x}_0, t_0,  \textbf{y}(\textbf{x}_0, t_0))_j\}_{j=1}^N$. Below we discuss several key aspects of data collection and learning.

\textbf{Data collection}: 
Finding BVP solutions with the existence of value discontinuity due to the collision requires softening the collision loss. This is done by using a sigmoid function and choosing its shape parameter that allows BVPs to be successfully solved. We solve the resultant BVP problems where the initial states are sampled from $\mathcal{X}_0$ and initial time $t=0$. Due to the softening, the resultant co-states are of the form
\begin{equation*}
    \lambda_i^*(t) = \exp\left(A_i^T (T-t)\right)\lambda_i^*(T) - q'_i \mathds{1}(t\geq t_i^{in}) + q'_i \mathds{1}(t \geq t_i^{out})
\end{equation*}
where $q'_i$ for non-collision close calls are estimated from the trajectory through least square during a post-process step.

\vspace{-0.05in}
\begin{algorithm}[h]
\SetAlgoLined
\SetKwInOut{Input}{input}
\SetKwInOut{Output}{output}
\Input{$\mathcal{X}_t$, $\mathcal{X}_a$}
\Output{$\hat{\textbf{y}}_w(\cdot)$}
    set $k=0$\;
    train $\hat{\textbf{y}}_w(\cdot)$ using $\mathcal{X}_t$\;
    \While{$\mathcal{X}_a \neq \emptyset$}{
        sample $\tilde{\mathcal{X}}_0^{(k)}  \subset \mathcal{X}_a$\;
        $\mathcal{A} = \emptyset$\;
        \For{$\textbf{x}_0 \in \tilde{\mathcal{X}}_0^{(k)}$ and $t_0 \in \mathcal{T}_0$}{
            compute $\hat{\boldsymbol{\lambda}}_{(\textbf{x}_0,t_0)}$ via $\hat{\textbf{y}}_w(\cdot)$\;
            compute $\hat{\textbf{u}}^*$ and $\hat{\textbf{x}}^*$ using $\hat{\boldsymbol{\lambda}}_{(\textbf{x}_0,t_0)}$ via Eq.~\eqref{eq:systemdynamics} and Eq.~\eqref{eq:costate_intersection}\;
            compute $\tilde{\boldsymbol{\lambda}}_{(\textbf{x}_0,t_0)}$ as solution to an inverse value problem with terminal values $\hat{\textbf{x}}^*(T)$ and $\hat{\boldsymbol{\lambda}}_{(\textbf{x}_0,t_0)}(T)$, and dynamics following Eq.~\eqref{eq:pmp}\;
            compute $a(\textbf{x}_0, t_0;w)$ via Eq.~\eqref{eq:acquisition}\;
            add $a(\textbf{x}_0, t_0;w)$ to $\mathcal{A}$\;
        }
        pick 
        $\textbf{x}^*$ and $t^*$ corresponding to $\max \mathcal{A}$\;
        compute $\textbf{y}(\textbf{x}^*, t^*)$ by solving the corresponding BVP\;
        add $(\textbf{x}^*, t^*, \textbf{y}(\textbf{x}^*, t^*))$ to $\mathcal{X}_t$\;
        remove $(\textbf{x}^*, t^*)$ from $\mathcal{X}_a$\;
        $k = k+1$\;
    }
\caption{Active learning}
\label{alg:active_learning}
\end{algorithm}
\vspace{-0.05in}

\textbf{Active learning}: Let the predicted co-state trajectory be $\hat{\boldsymbol{\lambda}}_{(\textbf{x}_0, t_0)}$. Here the dependence on $\textbf{x}_0$ and $t_0$ suggests that the prediction is derived from $\hat{\textbf{y}}_w(\textbf{x}_0, t_0)$. Using $\hat{\boldsymbol{\lambda}}_{(\textbf{x}_0,t_0)}$ and Eq.~\eqref{eq:pmp} we can compute $\hat{\textbf{u}}^*$ and then $\hat{\textbf{x}}^*$, which allows computing the co-state trajectory again through an inverse value problem. Denote this updated co-state trajectory by $\tilde{\boldsymbol{\lambda}}_{(\textbf{x}_0,t_0)}$, we can now introduce the following acquisition function:
\begin{equation}
    a(\textbf{x}_0, t_0;w) =  || \hat{\boldsymbol{\lambda}}_{(\textbf{x}_0,t_0)} - \tilde{\boldsymbol{\lambda}}_{(\textbf{x}_0,t_0)}||_2.
    \label{eq:acquisition}
\end{equation}
$a(\cdot;w)=0$ when $\hat{\textbf{y}}_w(\cdot)$ is fully generalizable. At the $k$th iteration of active learning, we sample a subset $\tilde{\mathcal{X}}_0^{(k)} \subset \mathcal{X}_0$, and solve the game $\mathcal{G}(\textbf{x}^*, t^*)$ for $\textbf{x}^*, t^*=\argmax_{\textbf{x} \in \tilde{\mathcal{X}}_0^{(k)}, t\in \mathcal{T}_0} a(\textbf{x},t;w)$. The resultant solution is then added to the dataset to update $w$. Alg.~\ref{alg:active_learning} summarizes the active learning algorithm, where $\mathcal{X}_t,~ \mathcal{X}_a \subset \mathcal{X}_0$ are the training and acquisition datasets, respectively. $\mathcal{X}_t \cap \mathcal{X}_a = \emptyset$.

\vspace{10pt}
\section{Case Study}

\subsection{Case setup}
Using the running example on the uncontrolled intersection, we learn three neural network models: one for value approximation, one for co-state approximation, and the last also for co-state approximation but through active learning. We use a variety of training data sizes for learning to observe how generalization performance improves as data increases. For each learning task, we perform twenty independent experiments to accommodate for randomness in training data sampling, model initialization, and the active learning algorithm. We consider two metrics for generalization performance: (1) the absolute prediction error in co-state at current time, and (2) the total number of collisions over a set of test cases by using the predictive model for closed-loop control. To remove confounding effects, the true equilibria of these test cases have no collisions. The second metric is of higher importance as it directly reflects the risk induced by the learned models.  

\textbf{Hypotheses}: From the experiments, we statistically test the following hypotheses: \textit{With the same training data sizes, closed-loop control using an actively learned co-state network leads to lower collision probability than using a statically learned co-state network, which in turn has lower collision probability than using a value network.}

\textbf{Data preparation}: The initial state set $\mathcal{X}_0$ is defined based on player position range $d_i \in [15, 20]m$ and velocity range $v_i \in [18, 25]m/s$. $\mathcal{T}_0$ is defined as $[0,3]$s with a time interval of $0.1$s. The training, acquisition, and test datasets (denoted by $\mathcal{X}_t$, $\mathcal{X}_a$, and $\mathcal{X}_e$, respectively) are independently sampled from $\mathcal{X}_0$ using a Latin Hypercube sampler with respective sizes 1573, 1000, and 575. For each sample initial state, the Nash equilibrial policies, values, and co-states are solved by the BVP solver for $t=0$. The resultant trajectories are interpolated over $\mathcal{T}_0$. Softening of the collision penalty is necessary for convergence, as described below. Lastly, we use multiple initial policy guesses to explore local equilibria, and choose the ones with the best sum of values across the players.

\textbf{Soften collision penalty for BVP convergence}: Following \cite{chen2020shall}, the collision penalty is modeled as
\begin{equation*}
    \phi_i(x) = b\sigma(d_i, \theta)\sigma(d_{-i}, 1),
\end{equation*}
where 
{\small
\begin{equation}
\begin{aligned}
    \sigma(d,\theta) = & \left(1+\exp(-\gamma (d-R/2+\theta W/2))\right)^{-1}\\
    & \left(1+\exp(\gamma (d-(R+W)/2-L))\right)^{-1},
\end{aligned}
\end{equation}
}$b=10^4$ sets a high loss for collision; $\gamma=5$ is a shape parameter empirically set to enable effective BVP convergence; $\theta=5$ is a parameter that sets a safety buffer zone to avoid actual collisions at the intersection. $R$, $L$, and $W$ are the road length, car length, and car width, respectively. These parameter settings and the visualization of collision penalty $\phi_i(x)$ are illustrated in Fig.~\ref{fig:case}.

\begin{figure}
    \centering
    \includegraphics[width = \linewidth]{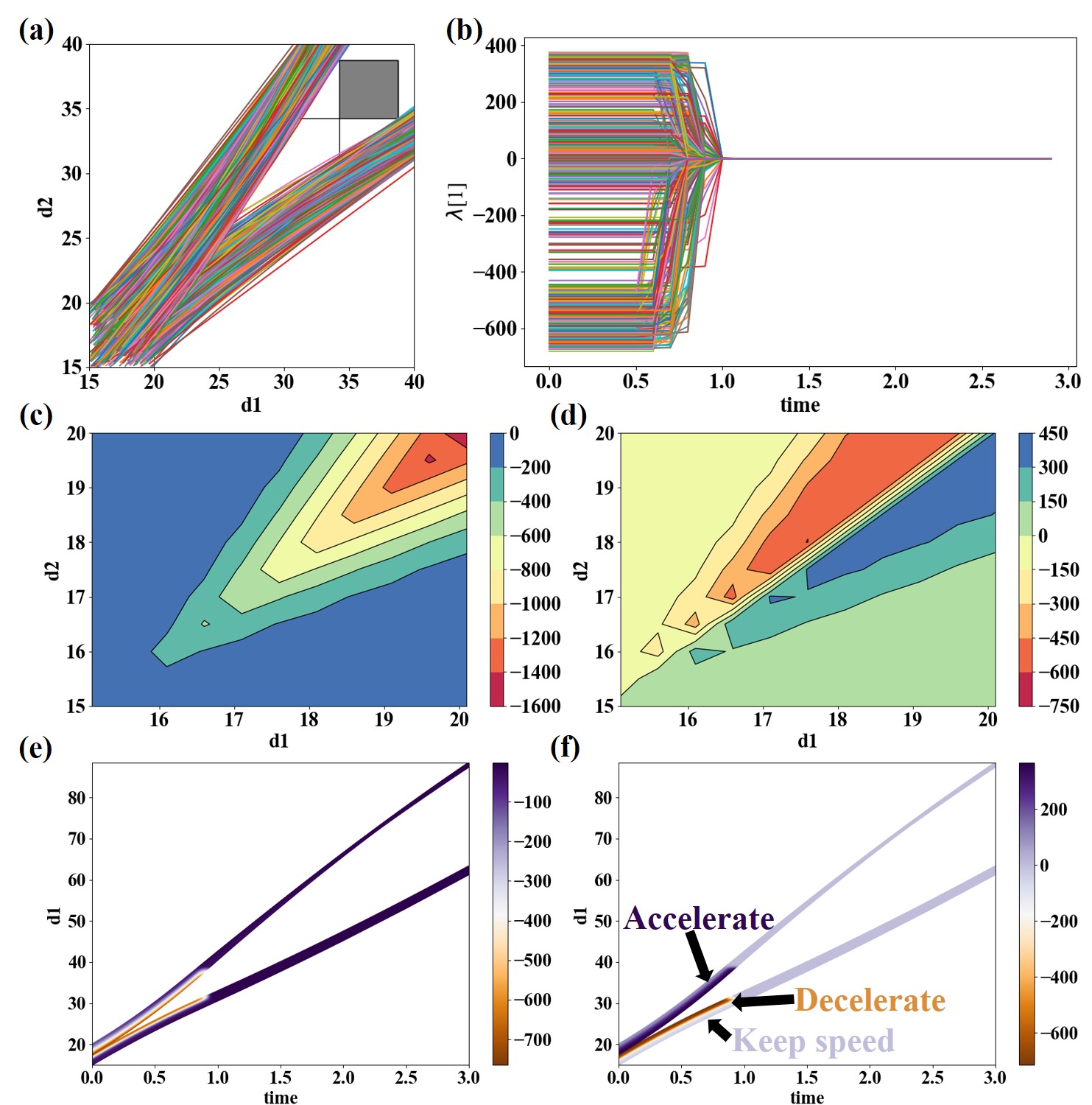}
    \caption{(a) Trajectories in the space of $d_1$ and $d_2$ driven by the equilibrium policies starting from different initial states. (b) The corresponding co-state trajectories along time. (c) Equilibrium value landscape across $d_1$ and $d_2$ with $t_0=0$ and $v_1=v_2=18m/s$. (d) Equilibrium co-state parameter $q_1$ landscape under the same settings of (c). (e, f) Equilibrium values and $q_1$ in the space of $d_1$ and $t$, where $d_2=17.5m$ and $v_1=v_2=18m/s$.}
    \label{fig:solution}
\vspace{-18pt}
\end{figure}

\textbf{Equilibrium solutions}: Fig.~\ref{fig:solution}a,b shows equilibrium policies and co-states sampled from $\mathcal{X}_0$. Consistent with our analysis, co-state trajectories have a low-dimensional representation, i.e., the trajectories over time can be compressed as four parameters. More specifically, each pair of co-state trajectories of Player 1 and 2 specifies: (1) the type of equilibrium interaction that is about to happen (see Fig. 2a of \cite{chen2020shall}) through $t_i^{in}$ and $t_i^{out}$, and (2) the risk of the interaction through $q_i$, where $q_i$ also directly governs the control policy according to Eq.~\eqref{eq:costate_intersection}. It is worth investigating in future studies whether this representation of co-states more closely reflects what human beings learn and use to make decision during collision-avoiding interactions.

We use Fig.~\ref{fig:solution}c-f to qualitatively explain why learning equilibrium co-states is more efficient than learning values. Our conjecture is that the equilibrium value landscape in the joint space of $\textbf{x}_0$ and $t_0$ has larger total variance than the equilibrium co-state landscapes. This is evident, for example, from the sample $d_1-t$ slice in Fig.~\ref{fig:solution}e,f where the value changes rapidly along $d_1$ due to the appearance of near-collision cases, while $q_1$ changes categorically from accelerating of Player 1 (when it is ahead of Player 2), to decelerating (when it is behind Player 2), and to keeping speed (when it is far away). Therefore, while learning of values requires capturing the highly nonlinear value variations, the learning of co-states boils down to a classification. For completeness, we also show landscapes of values of Player 1 and $q_1$ across $d_1$ and $d_2$ for fixed time and speeds in Fig.~\ref{fig:solution}c,d, respectively.

\textbf{Learning architecture and hyperparameters}:
From the value landscape in Fig. 2b of \cite{chen2020shall}, the equilibrium value exhibits two modes along time. Therefore we propose the following value network architecture:
\begin{equation}
    \hat{V}(\textbf{x},t) = \eta \mu(\textbf{x},t) + (1-\eta) \nu(\textbf{x},t),
\end{equation}
where $\mu$ and $\nu$ follow the same architecture: \texttt{fc5-(fc16-tanh)$\times$3-(fc2-tanh)}, and \texttt{fc}$n$ represents a fully connected layer with $n$ nodes and \texttt{tanh} is the hyperbolic tangent activation. $\eta$ is a sigmoid function that determines whether one of the agents have passed the intersection zone.

For co-state approximation, we adopt the same architecture as for $\mu$, yet expanding the output dimension of the last layer to match with that of the co-state parameters. It should be noted that a value network trained solely on $\mu$ failed to generalize. Therefore our choice of architecture is in favor of the value network in terms of model flexibility.

For each learning task, we perform full batch training use a learning rate of 0.01. For early termination, we set aside 400 independently drawn validation data points, along with their BVP solutions. The training terminates when it accumulates 500 epochs where the validation error, i.e., the $l_2$-norm between predicted and true values or co-state trajectories, increases.

\textbf{Algorithmic settings for active learning}: For each iteration of active learning, we uniformly sample 50 initial states from $\mathcal{X}_a$ and choose 10 with the highest $a(\cdot,w)$ values. An ablation study is yet to be performed to better understand the tradeoff between effectiveness and computational cost of active learning with respect to its hyperparameters.

\textbf{Closed-loop control}: For both value and co-state networks, we evaluate their generalization performance by counting their collision percentages among test cases when used for closed-loop control for both players. For the value network, we use PyTorch autograd to compute the co-state as $\nabla_{x_i}\hat{V}$. Following Eq.~\eqref{eq:costate_intersection}, the control input is then either $0.5\hat{\lambda}_i[2]$ when it is within the input bounds, or otherwise one of the bounds. We roll out the simulation using a time interval of 0.03s.  

\subsection{Results}
\begin{figure*}
  \includegraphics[width=\textwidth]{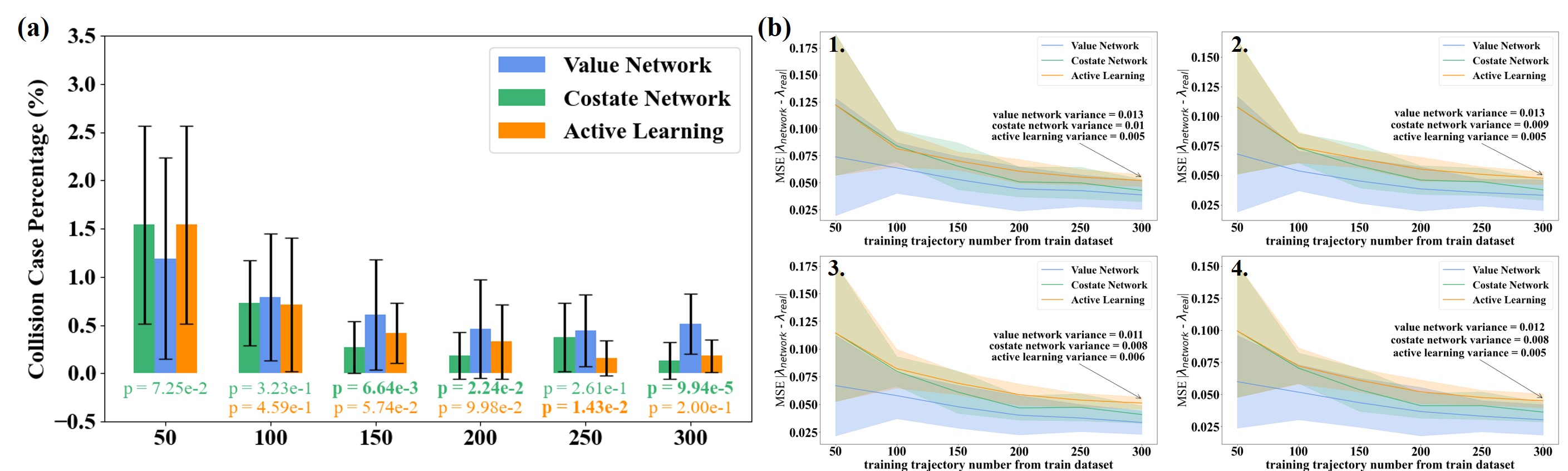}
  \caption{(a) Comparison on collision probabilities through closed-loop control by using the value network (blue), statically learned co-state network (green), and actively learned co-state network (orange). The p-values (green and orange) are for one-side t-tests between value and costate, and between active and static, respectively. Statistically significant p-values are highlighted. (b) Generalization performance by the three models on co-state prediction across training data sizes. 1-2: Co-state prediction errors for player 1; 3-4: for player 2.}
  \label{fig:result1}
\vspace{-15pt}
\end{figure*}

\begin{figure}
    \centering
    \includegraphics[width=\linewidth]{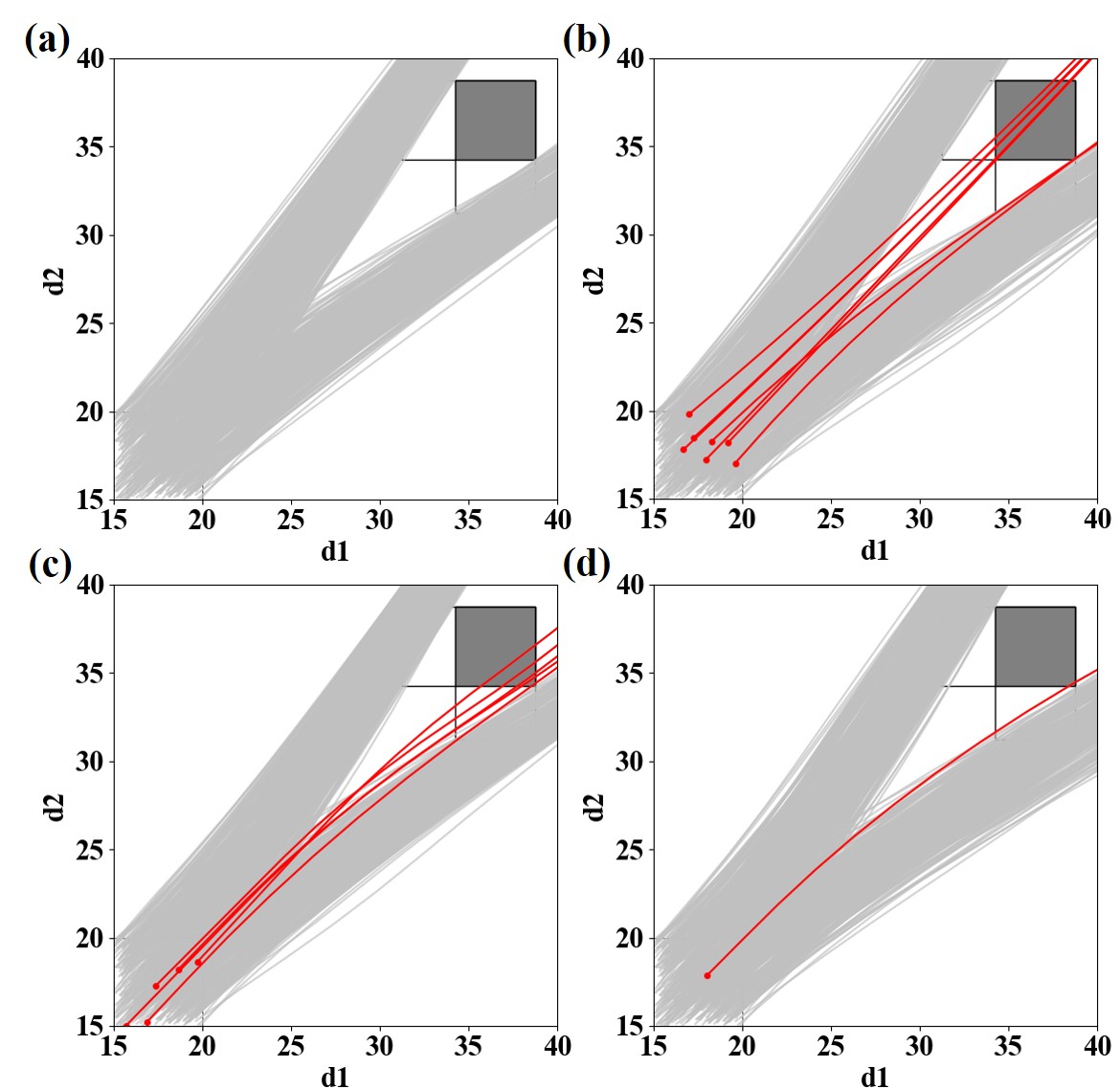}
    \caption{Test interaction trajectories from (a) the ground truth (BVP solver), and closed-loop control based on (b) the value network, (c) the statically trained co-state network, and (d) the actively trained co-state network. Training data size is 250. The grey box is the collision zone.}
    \label{fig:demo}
 \vspace{-0.3in}
\end{figure}

Fig.~\ref{fig:result1}a compares the collision percentages among all test cases contributed by the value and the co-state networks along increasing training data sizes. We perform t-tests to show statistically significant safety improvements contributed by co-state networks. We show in Fig.~\ref{fig:result1}b the mean squared errors of co-state predictions by the two networks. It is worth noting that while the performance between the two is statistically indifferent using this metric, the value network produces higher variances in its predictions under the same random procedure for sampling training data. We conjecture that the prediction variance correlates positively with the collision percentage. An inspection into this conjecture will be of interest for a future study. Fig.~\ref{fig:demo} demonstrates the test interaction trajectories along $d_1$ and $d_2$ from the ground truth, the value, and the co-state networks when the training data size is 250. Collision cases are highlighted. 

Lastly, our experiments show that the collision percentages of a statically and an actively trained co-state network are only statistically significant at an intermediate data size (250) (Fig.~\ref{fig:result1}a). An explanation for this is that Latin Hypercube sampling is by itself a theoretically efficient method, albeit static, and there is indeed a lack of proof on the advantage of active learning against Latin Hypercube sampling when the training and test data distributions are the same. In addition, when data size is sufficiently large (300 in this case), the advantage of active learning diminishes. 

\vspace{10pt}
\section{Discussion}
Here we summarize a few limitations of the presented work and feasible paths towards addressing them. \textbf{From linear to nonlinear dynamics}: We showed that equilibrium co-states have low-dimensional embedding when the system dynamics are linear. We first note that within the scope of autonomous driving, focused interaction problems, e.g., uncontrolled intersections, roundabout, and lane changing, can be approximated to have linear dynamics, in some cases under a Frenet coordinate system. In a more general scope, it is of interest to investigate whether such co-state structures can be obtained when approximated linear dynamics are learned from nonlinear systems, e.g., through Koopman operator theory~\cite{brunton2021modern}. \textbf{From two to multiple players}: The co-state structure we exploited is also contributed by the fact that there are only two players in the game, see Fig.~\ref{fig:discussion}a for an explanation. We conjecture that similar structures can be obtained from multi-player games when certain constraints are applied to the feasible inputs of players. For example, in an uncontrolled intersection case, the types of collisions can be enumerated, and thus the actual dimensionality of co-states can be determined, if no players can move backwards. See Fig.~\ref{fig:discussion}b for an example with three players.

\section{Conclusions}
This paper is concerned about generalization performance of data-driven Nash equilibrial policies in two-player differential games. Policies of such games follow HJI, and are known to be hard to compute directly, and data-intensive when approximated through learning.
We showed analytically that for collision-avoiding games with linear dynamics, the equilibrium co-states are low-dimensional due to finite equilibrium types. We then empirically showed the co-state networks lead to lower collision probabilities than value networks when trained on the same datasets. We also proposed a theory-driven active learning method where the quality of co-state prediction is evaluated by its compliance with PMP. We showed that active learning further improves collision probabilities from static learning when data size increases. Mathematically, we proposed to learn the manifold of dual variables (co-states) when it has a simple structure. Our future work will investigate the utility of this idea in multi-agent collision-avoiding interactions.

\section{Acknowledgments}
This work was in part supported by NSF CMMI-1925403 and NSF CNS-2101052. The views and conclusions contained in this document are those of the authors and should not be interpreted as representing the official policies, either expressed or implied, of the National Science Foundation or the U.S. Government.






\bibliographystyle{ieeetr}
\bibliography{acc}
\end{document}